# A Complementary Tracking Model with Multiple Features

Peng Gao, Yipeng Ma, Chao Li, Ke Song, Fei Wang, Liyi Xiao
Shenzhen Graduate School, Harbin Institute of Technology, China

***Abstract*—Discriminative Correlation Filters (DCF)-based tracking algorithms exploiting conventional handcrafted features have achieved impressive results both in terms of accuracy and robustness. Template handcrafted features have shown excellent performance, but they perform poorly when the appearance of target changes rapidly such as fast motions and fast deformations. In contrast, statistical handcrafted features are insensitive to fast states changes, but they yield inferior performance in the scenarios of illumination variations and background clutters. In this work, to achieve an efficient tracking performance, we propose a novel visual tracking algorithm, named MFCMT, based on a complementary ensemble model with multiple features, including Histogram of Oriented Gradients (HOGs), Color Names (CNs) and Color Histograms (CHs). Additionally, to improve tracking results and prevent targets drift, we introduce an effective fusion method by exploiting relative entropy to coalesce all basic response maps and get an optimal response. Furthermore, we suggest a simple but efficient update strategy to boost tracking performance. Comprehensive evaluations are conducted on two tracking benchmarks demonstrate and the experimental results demonstrate that our method is competitive with numerous state-of-the-art trackers. Our tracker achieves impressive performance with faster speed on these benchmarks.**

## I. Introduction

Visual object tracking is a basic research problem with various applications in computer vision. The goal of object tracking is to estimate the states of an arbitrary target by discriminating between its appearance and that of the surroundings in a video. Visual tracking can be applied in many applications, such as video surveillance, assistant driving systems and intelligent traffic control. The most effective trackers should handle all the variations both from background and target itself while track at speeds that far exceed the frame-rate requirement [1]. Despite much progress have been made in recent years, visual object tracking still remains largely unsolved problems due to various challenging factors such as fast motions, background clutters, motion blurs, deformations, illumination variations, in-plane rotations, low resolution, occlusions, out-of-plane rotations, out of views, scale variations, etc [2], [3].

In the past decades, DCF-based approaches [4]–[7] have drawn a lot of attention from the computer vision community because they are treated as similarity measurements between two image signals in signal processing. Due to DCF enables training and detection with dense sampling strategy and high dimensional features [4], [5] at frame-rate by efficiently solving a ridge regression problem in the Fourier frequency domain, the time-consuming convolution operations can be straightforwardly avoided. Most state-of-the-art DCF-based trackers employ template handcrafted features such as HOGs and CNs to present the target in a video, they have shown excellent performance on existing visual object tracking benchmarks, but they perform poorly when the appearance of the target changes rapidly such as fast motions and fast deformations. In contrast, statistical handcrafted features, such as CH, are insensitive to fast target states changing, but they yield inferior performance when illumination variations and background clutters are considered [8]. It is proved that exploiting powerful object representations can significantly improve tracking performance [9], some recent tracking algorithms [10]–[12] concentrate on exploiting all the object representations above-mentioned to train a model, which show favorable performance to target states change and color variations. However, these methods combine the response maps simply, the tracking results are not the optimal.

Motivated by these facts, we propose an appealing complementary ensemble object tracking framework to take advantage of multiple features for visual object tracking. Specifically, we deduce several basic trackers by treating multiple features as independent linear regression problems. Each basic tracker gives a response map of the target position in a new image. Moreover, we observe that the response maps obtained by a single basic tracker are not robust enough to handle more challenge scenarios. Therefore, a continuous fusion technique is proposed, which can significantly improve tracking performance by coalescing the basic response maps using relative entropy, then we can yield more robust and reliable tracking results. Additionally, to further strengthen the proposed tracker to deal with more challenge scenarios, we exploit a robust model update strategy for our tracker. Comprehensive evaluations demonstrate that our proposed complementary ensemble tracking algorithm achieves a considerable performance improvement compared with a variety of state-of-the-art approaches. Since template response maps are carried out in the Fourier frequency domain, and statistical response map is calculated by resizing the image path to a restricted size in advance, these enable our tracker to run at speeds that far exceed the frame-rate requirement.

The rest of the paper is organized as follows. Section 2 briefly reviews some related works. Section 3 details our proposed tracking algorithm. Section 4 demonstrates the implementation details and experimental results. Finally, Section 5 draws a conclusion of our work.

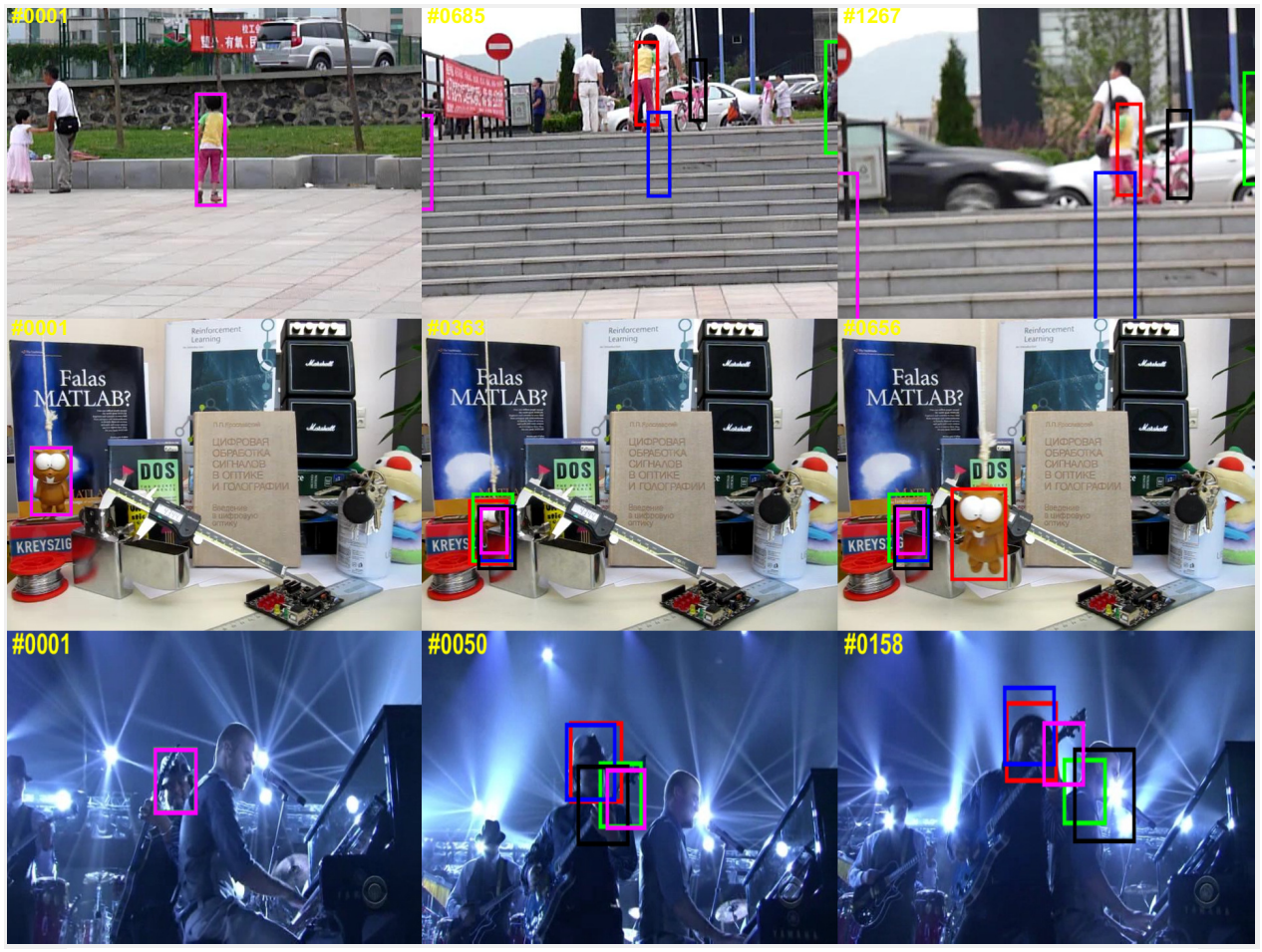

Fig. 1: A comparison of our method MFCMT with four state-of-the-art trackers Staple [10], MEEM [12], DSST [6] and KCF [5] on three example sequences *Girl2*, *Lemming* and *Shaking* in OTB [2], [3], respectively. Best viewed in color.

## II. Related works

### A. Template features-based trackers

DCF-based tracking algorithms have been ignited by the impressive success of MOSSE [13] in visual tracking field since it employs the raw pixel and classical signal processing analysis to obtain fast correlation filters and the significant computational efficiency. CSK [4] is one of the subsequent algorithms, it explores a circular structure of patch to improve the classify ability, which employs kernelized correlation filters to achieve high efficiency. KCF [5] and CN [14] incorporate multi-channel conventional handcrafted features to enhance the performance of CSK. However, KCF and CN employ fixed-sized patches during tracking, which is not able to deal with scale variations of targets. To resolve this issue, more works have been proposed. DSST [6] introduce an efficient method for estimating the target's scale by training classifiers on a scale pyramid. Meanwhile, DCF-based trackers perform poorly when the appearance of objects changes rapidly such as fast motions and fast deformations [2], [3].

### B. Statistical features-based trackers

As a kind of earlier feature representation for tracking, statistical features, especially CH, are used to present the appearance of the targets [15]. Since statistical features are insensitive to fast target state changes and cannot be affected by the boundary effects, they can be considered orthogonal to the template features. But such models may drift to adjacent regions that represent a similar appearance compared to the target region. A canonical statistical features-based tracker is DAT [8] that adapt the object representation to suppress the regions with similar colors by using a naïve Bayes classifier.

### C. Multiple features-based trackers

It has been proven that the result of a single tracker may sometimes be unreliable, but the ensemble trackers which have high diversity always can improve the performance [9]. To achieve excellent results both in terms of accuracy and robustness, CSR-DCF [11] exploits a channel and spatial reliability scheme to adaptively learn and track the apparent objects color region. Nonetheless, it severely compromises the speed of the system. Staple [10] proposes a simple interpolated combination of multiple response scores in a dense search. However, Staple uses a fixed combination factor to linear interpolate the response maps, it seems to trust the response of correlation filters more than the color histograms. This combination strategy is too simple to gain robustness tracking results.

Staple [10] is most closely related to our work. However, different from Staple, we are carefully deriving an efficient complementary ensemble tracking framework to significantly improve the tracking performance. Moreover, we propose an effective fusion technique instead of using a fixed factor to linear interpolate the response maps in Staple, thus our tracker can yield more robust and reliable tracking results. We also suggest a robust model update strategy inspired to further enhance the proposed tracker to deal with more challenge scenarios with faster speed.

## III. The Proposed Approach

In this section, we first introduce the formulation of our complementary model with multiple features. Then, we present a robust ensemble approach aim to build a better observation model to reciprocally compensate the deficiencies of single primary trackers. Finally, we propose an online model update strategy to avoid model drifting and corrupting caused by similar objects or surrounding background appearance.

### A. Basic models

In our algorithm, we exploit DCF to construct the template features based basic model due to their lower computational burden. The key point of DCF-based trackers is that the high-dimensional augmentation of negative samples are employed as the train samples to enhance the discriminative ability of the tracking-by-detection framework [4], [5], [16]. Additionally, DCF can quickly find the linear model that provides the best fit to the desired correlation output in the least-squares sense. They are attractive algorithms for tracking, due to their excellent performance and high computational efficiency. We use CHs to construct the statistical features-based basic model. Fig.2 shows the framework of our method. For formulating the basic models, we consider them as several independent ridge regression problems due to the advantage of having a closed-form solution for the optimization problem, which makes it much easier to calculate.

We train several ridge regressions on the $L$-channel template feature maps $\mathbf{x}_{t,u} \in \mathbb{R}^{M \times N \times L}$ , including raw pixels, HOGs and CNs, in frame $u$, and the desired correlation output $\mathbf{y}_{t,u} \in \mathbb{R}^{M \times N}$ which typically follow a Gaussian function with peak value 1 at the center. Our aim is to find an optimal correlation filter $\mathbf{w}_u$. This can be obtained by minimizing a ridge regression in the Fourier domain as:

$$\min_{\mathbf{w}_u^l} \| \sum_{l=1}^{L} \mathbf{w}_u^l \star \mathbf{x}_{t,u}^l - \mathbf{y}_{t,u} \|^2 + \lambda \sum_{l=1}^{L} \|\mathbf{w}_u^l\|^2 \tag{1}$$

where $\mathbf{w}_u^l$ refers to the channel $l$ of the DCF $\mathbf{w}_u$ in the frame $u$, the star symbol $\star$ denotes circular cross-correlation, $\mathbf{x}_{t,u}^l$, $\mathbf{y}_{t,u}$ and $\mathbf{w}_u^l$ are all of size $M \times N$ . Following Parseval's theorem and adopting the properties of circulant matrix in CSK [4] and KCF [5], the solution can be gained as:

$$\mathcal{W}_u^l = \frac{\mathcal{Y}_{t,u}^* \odot \mathcal{X}_{t,u}^l}{\sum_{l=1}^{L} (\mathcal{X}_{t,u}^l)^* \odot \mathcal{X}_{t,u}^l + \lambda} \tag{2}$$

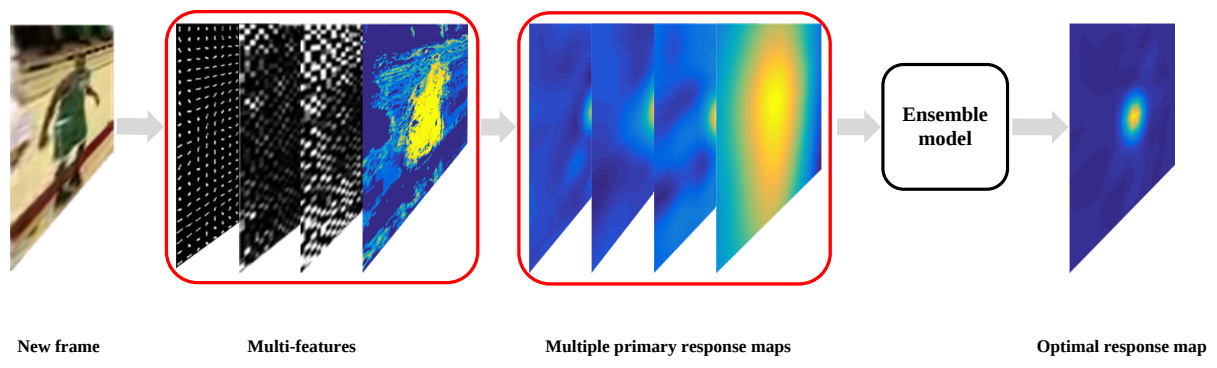

Fig. 2: The framework of our proposed method.

Here $\mathcal{W}_u = \mathcal{F}(\mathbf{w}_u)$, and $\mathcal{F}(\cdot)$ denotes the Discrete Fourier Transform (DFT). We using the symbol $*$ to represent the complex conjugation $\mathcal{Y}_u^*$ of a complex number $\mathcal{Y}_u$, the symbol $\odot$ for element-wise multiplication, namely Hadamard product, and $\div$ denotes the element-wise division. Additionally, the regularization coefficient $\lambda$ alleviates division-by-zero.

During detection, given the template features $\mathbf{z}_{t,u+1}$ which are extracted from a basis sample patch centred at the previous position of the target in a new frame $(u+1)$, the response map can be obtained by the following cross-correlation formulation:

$$\tilde{\mathbf{y}}_{t,u+1} = \mathcal{F}^{-1}(\sum_{l=1}^{L} \mathcal{W}_s^l \odot \mathcal{Z}_{t,u+1}^l) \tag{3}$$

where $\tilde{\mathbf{y}}_{t,u+1} \in \mathbb{R}^{M \times N}$ is the response map of predicted states in the new frame, and $\mathcal{F}^{-1}(\cdot)$ denotes the Inverse Discrete Fourier Transform (IDFT).

We also apply a ridge regression to the statistical features $\mathbf{x}_{s,u} \in \mathbb{R}^{M \times N \times K}$ over the foreground and background regions $\mathrm{F}_u$ and $\mathrm{B}_u$ cropped from an input image patch $\mathrm{I}_u \in \mathbb{R}^{M \times N}$ independently. This can be obtained by minimizing a ridge regression problem as:

$$\min_{\mathbf{v}_u} \sum_{\mathbf{x}_{s,u} \in \mathrm{I}_u} \|\mathbf{v}_u^T \mathbf{x}_{s,u} - \mathbf{y}_{s,u}\|^2 \tag{4}$$

Here, $\mathbf{y}_{s,u}$ is the corresponding regression, i.e. $\mathbf{y}_{s,u} = 1$ for positive samples or $\mathbf{y}_{s,u} = 0$ for negative samples, the parameter vector $\mathbf{v}_u$ can be learned from the color histograms of the previous frames. We decompose the regression into two independent terms of each region

$$\min_{\mathbf{v}_u} \sum_{\mathbf{x}_{s,u} \in \mathrm{F}_u} \|\mathbf{v}_u^T \mathbf{x}_{s,u} - 1\|^2 + \sum_{\mathbf{x}_{s,u} \in \mathrm{B}_u} \|\mathbf{v}_u^T \mathbf{x}_{s,u} - 0\|^2 \tag{5}$$

Thus, the sparse inner product is simply a lookup matrix that $\mathbf{v}_u^T \mathbf{x}_{s,u} = \mathbf{v}_u^{\mathbf{x}_{s,u}}$, where $\mathbf{v}_u^{\mathbf{x}_{s,u}}$ indicates the element of $\mathbf{v}_u$ for which the channel index is non-zero. Then, we adopt color histogram $\mathcal{H}(\mathbf{x}_{s,u})$ to present the object pixels, the solution can be obtained as

$$\mathbf{v}_{u+1} = \begin{cases} \frac{\mathcal{H}_{\mathrm{F}_u}(\mathbf{x}_{s,u})}{\mathcal{H}_{\mathrm{F}_u}(\mathbf{x}_{s,u}) + \mathcal{H}_{\mathrm{B}_u}(\mathbf{x}_{s,u})} & \text{if } \mathbf{x}_{s,u} \in \mathrm{F}_u \cup \mathrm{B}_u \\ 0 & \text{otherwise} \end{cases} \tag{6}$$

Since $\mathbf{x}_{s,u}$ is sparse, the statistical features-based basic model can be computed over a large detection region and a very low computational cost.

Finally, it can be sped up for the detection process by adopting the integral image method [17] to obtain the statistical response map

$$\begin{aligned} \tilde{\mathbf{y}}_{s,u+1}(i,j) =& \mathbf{v}_{u+1}(i,j) + \tilde{\mathbf{y}}_{s,u+1}(i,j-1) + \\ & \tilde{\mathbf{y}}_{s,u+1}(i-1,j) - \tilde{\mathbf{y}}_{s,u+1}(i-1,j-1) \end{aligned} \tag{7}$$

where $\tilde{\mathbf{y}}_{s,u+1}(i,-1) = 0$ and $\tilde{\mathbf{y}}_{s,u+1}(-1,j) = 0$.

For the template feature map of size $M \times N \times L$, the complexity of computing the correlation filters is $\mathcal{O}(LMN \log MN)$ due to it only requires element-wise product, and the online update process requires $\mathcal{O}(LMN)$ memory. This reduces storage and computation by several orders of magnitude.

### B. *Ensemble model*

For visual tracking, response maps obtained by a single basic model are not robust to deal with more challenging scenarios. Therefore, a robust tracker requires multiple kinds of features. We propose an ensemble model for the visual object tracking that leverages multiple features, which can significantly improve tracking performance by coalescing response maps based on relative entropy, so we can yield more robust and reliable tracking results.

In our method, we select the multiple features from both the template and statistical handcrafted features, including raw pixels, HOGs, CNs and CHs. For the $i^{th}$ basic response maps $P_u = \{P^1, P^2, ..., P^i\}$, the single response map $P_u^l \in P_u$, $l = 1, 2, ..., i$, can be considered as a probability map, which consists of a probability distribution $p_{(m,n)}^l$, $(m,n) \in \{1, 2, ..., M\} \times \{1, 2, ..., N\}$. The probability distribution is subjected to $\sum p_{(m,n)}^l = 1$ and indicates the probability that position $(m,n)$ is the centroid of the predicted bounding box. Consequently, in order to find the optimal response map $Q_u$ of our ensemble model, we can minimize the relative entropy, i.e. Kullback-Leibler divergence, between each single response map $P_u^l$ and the optimal response map $Q_u$. Relative entropy can be calculated as:

$$KL(P_u^l \| Q_u) = \sum_{(m,n)} p_{(m,n)}^l \log \frac{p_{(m,n)}^l}{q_{(m,n)}} \tag{8}$$

Thus, we can obtain the optimal response map $Q_u$ by

$$\begin{aligned} &\arg\min_{Q_u} \sum_{l=1}^{i} KL(P_u^l \| Q_u) \\ &s.t. \sum q_{(m,n)} = 1 \end{aligned} \tag{9}$$

where $p_{(m,n)}$ and $q_{(m,n)}$ denote the $(m,n)^{th}$ elements of primary response maps $P_u$ and optimal response map $Q_u$, respectively.

To solve the above equation, we can exploit the Lagrange multiplier method.

Finally, we can get the centroid $(x_u, y_u)$ of the predicted bounding box at a new frame by find the maximum value of the optimal response map $Q_u$ of our ensemble model

$$(x_u, y_u) = \arg\max_{m,n} Q_u(m,n) \tag{10}$$

We can see that the optimal response map is obtained by the sum of all the primary response maps, which can enhance the final result by all the primary response maps from multiple handcraft features using a weighted sum.

### *C. Model update*

For the online update, we should update the models using the most reliable frames. We consider three criteria, the highest peak values of optimal response maps, template features response maps and statistic features response map, i.e. $\max(Q_u)$, $\max(\tilde{\mathbf{y}}_{t,u})$ and $\max(\tilde{\mathbf{y}}_{s,u})$, respectively. When all these three criteria of the current frame are great than their corresponding average values of the last 10 frames with predefined thresholds $\gamma_q$, $\gamma_t$ and $\gamma_s$, the detected result in the current frame is considered to be highly reliable. Then we can update the proposed ensemble model with learning rate parameters for different kind of features as

$$\begin{aligned} \mathcal{W}^u_{new} &= (1-\eta_t)\mathcal{W}^{u-1} + \eta_t \mathcal{W}^u \\ \mathcal{H}^u_{new} &= (1-\eta_s)\mathcal{H}^{u-1} + \eta_s \mathcal{H}^u \end{aligned} \tag{11}$$

where $\eta_t$ and $\eta_s$ are the learning rate which set to 0.02 and 0.04, respectively.

Additionally, we first estimate the position, then search on the scale similar to DSST [6] by adopting a multi-scale template model at the estimated object position to handle object variation. And the multi-scale template model only executed when the full response maps are more reliable. The whole procedure is summarized in Algorithm 1.

## IV. Experiments

We perform all the experiments of our proposed tracker on OTB-2013 and OTB-2015 benchmarks [2], [3] which contain 51 sequences and 100 sequences, respectively. The results demonstrate that our algorithm achieves considerable tracking results.

### *A. Implementation details*

We exploit raw pixels, HOGs and CNs as template features to construct three different kinds of basic models. The cell size of HOGs is 4×4 and the orientation bin number of HOGs is 9, we also multiple the template features by a Hanning window during tracking. We exploit the distribution of color values in the RGB cube with histograms using 32 bins per channel in the statistic models, namely the color bin number of our color histograms is 32×32×32. For each frame, the search region is cropped twice the last object size

**Algorithm 1** The Proposed Approach (MFCMT)

**Input:** images $\{\mathbf{I}_u\}_{u=1}^U$, initial object position $\mathbf{p}_1$
**Output:** estimated object position of each frame $\{\mathbf{p}_u\}_{u=2}^U$
**for** $u = 2, \dots, U$ **do**
  Crop an image region $\mathrm{I}_u$ from $\mathbf{I}_u$ at the previous position $\mathbf{p}_{u-1}$
  Crop foreground and background regions $\mathrm{F}_u$ and $\mathrm{B}_u$ from $\mathrm{I}_u$
  Extract template and statistic feature maps $\mathbf{x}_{t,u}$ and $\mathbf{x}_{s,u}$
  Compute the template and statistic primary response maps $\tilde{\mathbf{y}}_{t,u}$ and $\tilde{\mathbf{y}}_{s,u}$ using Eq. 3 and Eq. 7
  Obtain the optimal response map $Q_u$ using Eq. 9
  Set $\mathbf{p}_u$ to the current object position at the higher peak value using Eq. 10
  Estimate the scale of object refer to [6]
  **if** $\max(Q_u)$, $\max(\tilde{\mathbf{y}}_{t,u})$ and $\max(\tilde{\mathbf{y}}_{s,u})$ satisfy the update conditions **then**
    Update the scale estimation model refer to [6]
    Update the model $\mathcal{W}_u$ and $\mathcal{H}_u$ using Eq. 11
  **end if**
**end for**

around the estimated position, and the cropped search region is resized to a fixed size 150×150. In order to compute the histograms more correctly, we define 85% of the last object size as the foreground surrounded by the background patch to avoid mislabeling as suggested in DAT [8]. The update predefined thresholds $\gamma_F$, $\gamma_T$ and $\gamma_S$ are set to 0.5, 0.7 and 0.5, respectively. The regularization coefficient in 2 is $\lambda_T = 0.001$. The standard deviation of the desired correlation response output is set to 1/16 of the object size. Similar to DSST [6], we use 33 number of scales with a scale of 1.02 in the scale estimation model.

The proposed tracker is implemented in MATLAB. All experiments are conducted on an Intel i5-4590 CPU at 3.3GHz with 8GB RAM.

### *B. Evaluation on OTB*

To evaluate our proposed tracker comprehensively, we follow the protocol of OTB [2], [3]. We use one-pass evaluation (OPE), temporal robustness evaluation (TRE) and spatial robustness evaluation (SRE) as suggested by OTB to verify the robustness of our tracker. We also report the speed of the trackers in average frames per second (FPS) over all the sequences. We present both the metrics mentioned above by precision and success plots. We provide a comprehensive comparison of our approach with several state-of-the-art trackers including Staple [10], SAMF [18], MEEM [12], DSST [6], DLSSVM [19], KCF [5] and Struck [20]. Here, Staple [10] can be considered as the simple version of our MFCMT that does not employ neither the proposed ensemble method or the model update strategy, and SAMF [18] is another variation of our method that without none of CHs ,the ensemble method or the model update strategy.

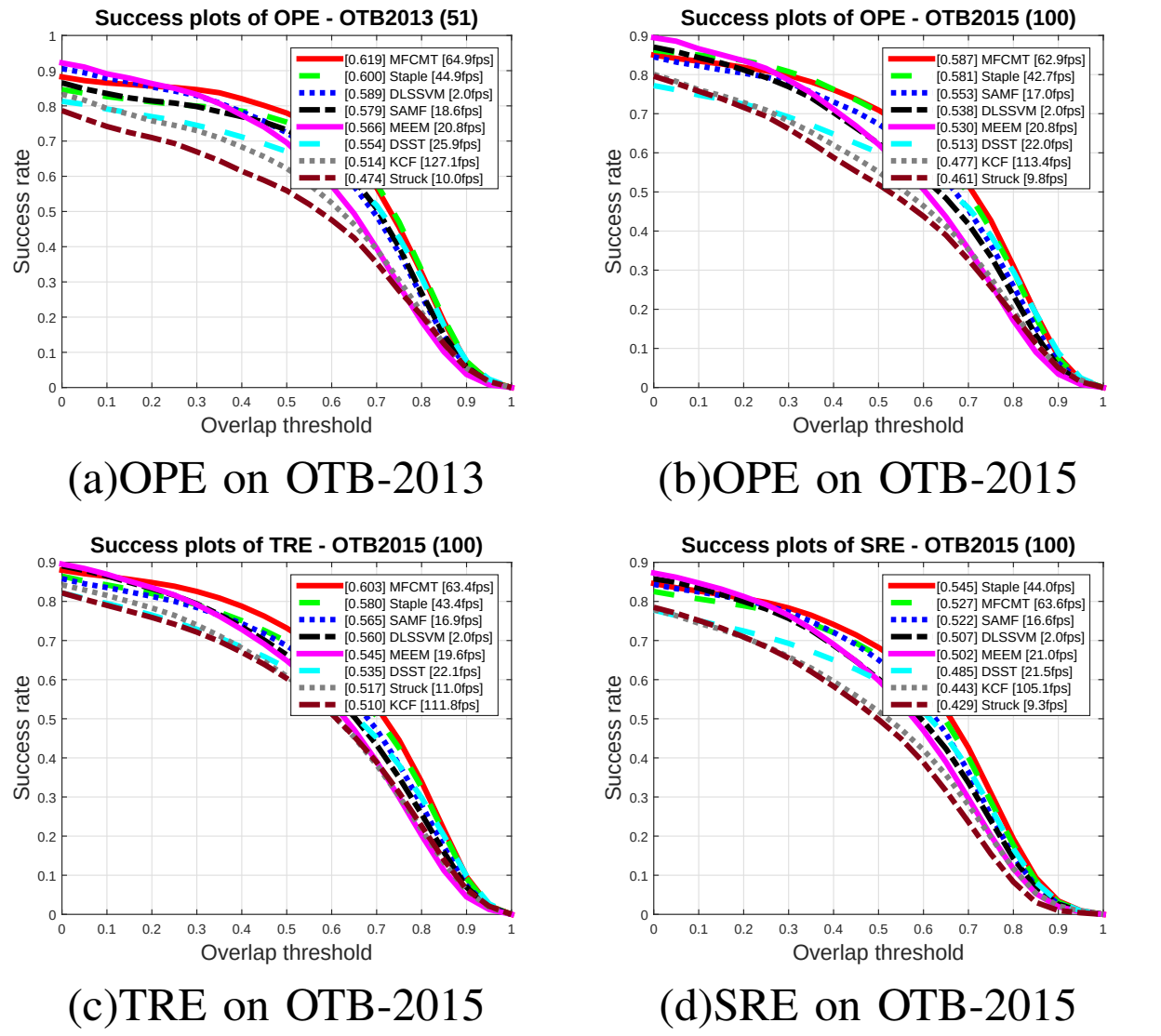

(a)OPE on OTB-2013 (b)OPE on OTB-2015

(c)TRE on OTB-2015 (d)SRE on OTB-2015

Fig. 3: Success plots of OPE, TRE and SRE for the top ten trackers in our comparison on the OTB [2], [3]. The first value in the legend indicate the scores for each tracker. The last numbers in the bracket denote the speed of the trackers in mean FPS. Best viewed in color.

Fig.3 illustrates the success plots of the evaluation participated trackers on both OTB-2013 and OTB-2015. Our complementary ensemble trackers (MFCMT) performs best with all evaluation metrics in the two benchmarks. The results of TRE and SRE show the robustness of our ensemble model. Our tracker leads to 27% gains in success plots on OTB-2015 compared with Struck which is the best performance tracker in original benchmark [2]. Staple and MEEM are trackers which developed based on multiple trackers, our tracker significantly improves them by 4% and 10%, respectively. KCF is the basis of our template features primary model, and DSST provides the scale estimation method, but our tracker performs favorably over them while by 20% and 13%, respectively. As for tracking speed, our tracker also run at a significantly high speed of 63 FPS, which superior to other up-to-date trackers, including MEEM, DLSSVM and Staple.

Additionally, the video sequences contained in both the OTB-2013 and OTB-2015 benchmarks are annotated with eleven different attributes, such as fast motion, background clutter, motion blur, deformation, illumination variation, in-plane rotation, low resolution, occlusion, out-of-plane rotation, out of view and scale variation. All these attributes represent various challenging factors. For detailed experiments, we also evaluate our method on various challenging attributes in OTB-2015. The complete comparison on 6 of 11 attributes are illustrated in Fig.4, showing the favorable performance of our ensemble method compared to all the participate state-of-the-art trackers for all attributes, include Staple [10], SAM-F [18], DSST [6], MEEM [12], KCF [5], DLSSVM [19],and Struck [20]. It is clear that our approach can deal with most challenging factors at speeds that far exceed the frame-rate requirement.

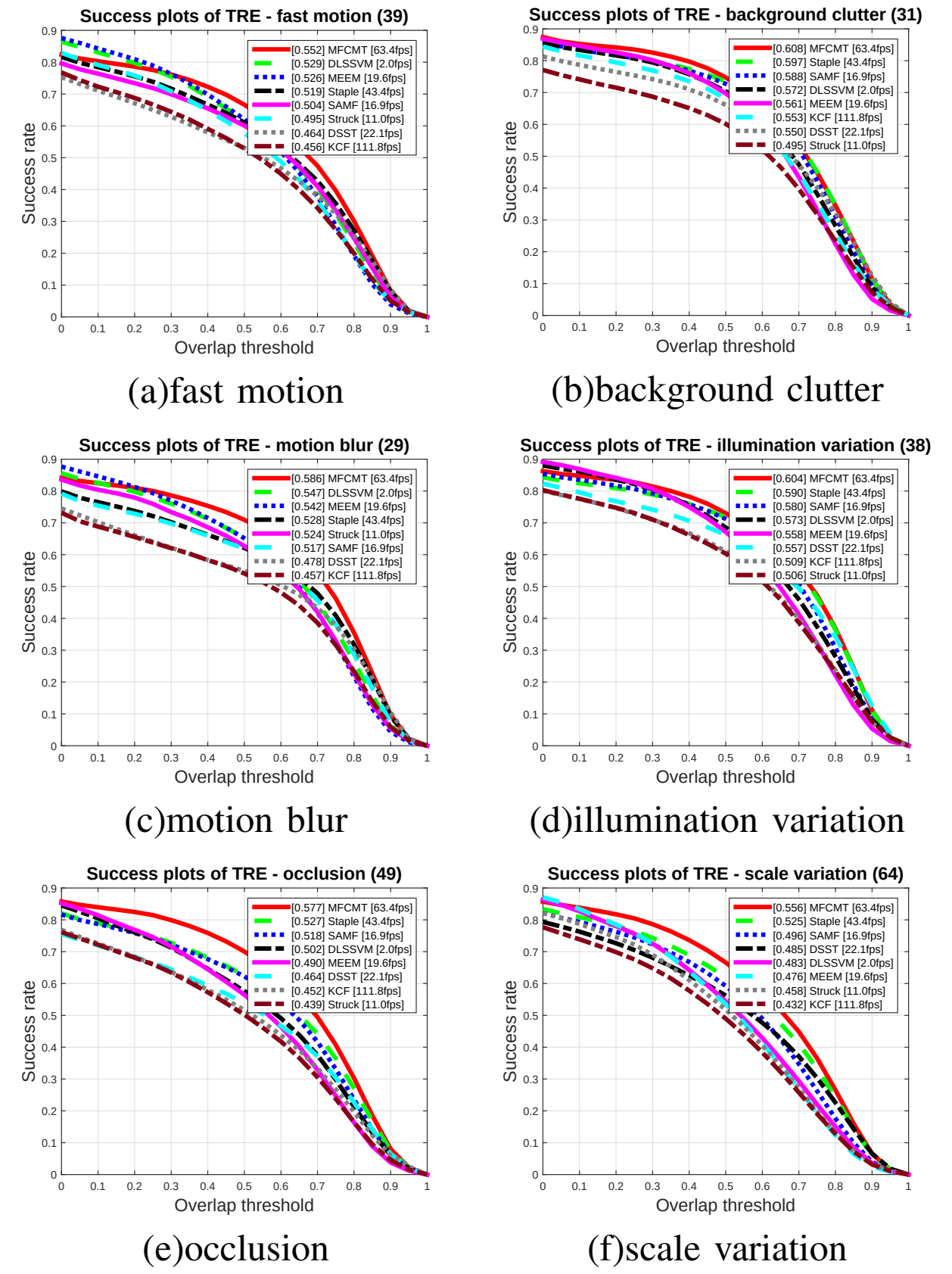

(a)fast motion (b)background clutter

(c)motion blur (d)illumination variation

(e)occlusion (f)scale variation

Fig. 4: The success plot of 6 attribute sub-datasets on OTB-2015. The value appears in the title is the number of sequences in the sub-dataset. The first value in the legend indicate the scores for each tracker. The last numbers in the bracket denote the speed of the trackers in mean FPS. Best viewed in color.

## V. Conclusions

In this paper, we propose a novel complementary ensemble object tracking method with multiple features. We observe that the response map obtained by a single model cannot deal with more challenges. Therefore, we adopt a relative entropy-based continuously fusion technique to further enhance the proposed tracker to deal with more challenge scenarios and improve tracking performance. Response maps obtained from multiple features-based models can be coalesced using our ensemble method. Furthermore, we also present a simple and effective model update strategy to improve both the accuracy and robust of tracking performance. We conduct evaluations on modern online tracking benchmarks. The evaluation results demonstrate that the proposed method is more effective and faster than several state-of-the-art trackers.